\documentclass{ieeeconf}

\usepackage{cite}
\usepackage{enumerate}
\usepackage{amsmath}
\usepackage{hyperref}
\usepackage{graphicx} 
\usepackage{xcolor}
\usepackage{amsmath}
\usepackage{amssymb}
\let\labelindent\relax
\usepackage{enumitem}
\usepackage{marvosym}

\hypersetup{
    colorlinks=true,          
    linkcolor=blue,           
    citecolor=green,          
    filecolor=magenta,        
    urlcolor=cyan,            
    pdfborder={0 0 1},        
    linkbordercolor={0 1 0},  
    citebordercolor={1 0 0},  
    filebordercolor={0 0 1},  
    urlbordercolor={0 1 1}    
}

\title{\LARGE \bf

Robo-MUTUAL: Robotic Multimodal Task Specification via Unimodal Learning
}

\author{Jianxiong Li$^{1*}$, Zhihao Wang$^{2, 1*}$, Jinliang Zheng$^{1, 3*}$, Xiaoai Zhou$^{4,1}$, Guanming Wang$^{5,1}$, Guanglu Song$^{3}$,\\ Yu Liu$^{3}$, Jingjing Liu$^{1}$, Ya-Qin Zhang$^{1}$, Junzhi Yu$^{2}$\textsuperscript{\Letter} and Xianyuan Zhan$^{1, 6}$\textsuperscript{\Letter}
\thanks{*Equal Contribution. 
$^{1}$Institute for AI Industry Research (AIR), Tsinghua University. $^{2}$College of Engineering, Peking University. $^{3}$Sensetime Research. $^{4}$University of Toronto. $^{5}$University College London. $^{6}$Shanghai AI Lab. \textsuperscript{\Letter} Correspondence to {\tt\small yujunzhi@pku.edu.cn, zhanxianyuan@air.tsinghua.edu.cn}.}
}

\begin{document}

\maketitle
\thispagestyle{empty}
\pagestyle{empty}

\begin{abstract}
Multimodal task specification is essential for enhanced robotic performance, where \textit{Cross-modality Alignment} enables the robot to holistically understand complex task instructions. Directly annotating multimodal instructions for model training proves impractical, due to the sparsity of paired multimodal data. In this study, we demonstrate that by leveraging unimodal instructions abundant in real data, we can effectively teach robots to learn multimodal task specifications. First, we endow the robot with strong \textit{Cross-modality Alignment} capabilities, by pretraining a  robotic multimodal encoder using extensive out-of-domain data. Then, we employ two Collapse and Corrupt operations to further bridge the remaining modality gap in the learned multimodal representation. This approach projects different modalities of identical task goal as interchangeable representations, thus enabling accurate robotic operations within a well-aligned multimodal latent space. Evaluation across more than 130 tasks and 4000 evaluations on both simulated LIBERO benchmark and real robot platforms showcases the superior capabilities of our proposed framework, demonstrating significant advantage in overcoming data constraints in robotic learning. Website: 
\href{zh1hao.wang/Robo_MUTUAL}{\texttt{zh1hao.wang/Robo\_MUTUAL}}
\end{abstract}


\section{INTRODUCTION}

Developing robots that can understand task specifications from diverse modalities (e.g., image, video, text, speech) is a pivotal research area in robot learning~\cite{shah2023mutex, reuss2024multimodal, octo_2023}. This not only enhances robot performance but also enriches the human-robot interaction experience~\cite{jiang2023vima, driess2023palme, myers2023goal, yu2023using}. 
To correctly interpret multimodal task specifications, one essential capability required is \textit{Cross-modality Alignment}~\cite{shah2023mutex,jiang2023vima,yu2023using, nguyen2021practical}, where an integral high-level task goal exists across various modalities of instructions (or known as prompts~\cite{shah2023mutex, jiang2023vima}) to prevent confusion. Existing methods typically acquire this ability through extensive end-to-end training, raising high demand for meticulously annotated multimodal prompts~\cite{octo_2023,shah2023mutex,reuss2024multimodal}. Collecting such prompts via crowd-sourcing is notably expensive and laborious~\cite{shah2023mutex, xiao2022dial}, not to mention impractical, when only unimodal prompts are accessible (\textit{e.g.}, text instructions are absent and only visual goals are provided). There have been attempts to synthetically generate missing prompts, but how to ensure data quality remains an open question~\cite{xiao2022dial}. Therefore, we wonder 
Can we bypass the stringent demands for paired multimodal prompts via unimodal task learning? overcoming significant data constraints and opening new avenues for efficient robots learning.

\begin{figure}
    \centering
    \includegraphics[width=1.0\linewidth]{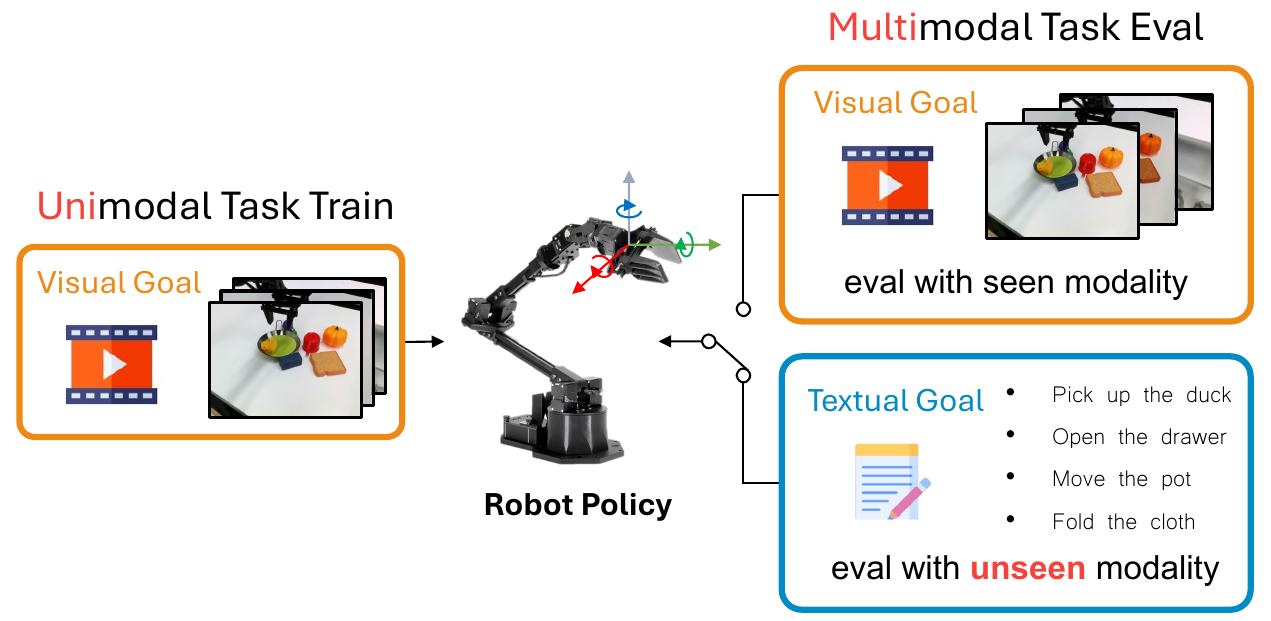}
    \vspace{-20pt}
    \caption{Training robot policies on unimodal task prompts but evaluate using prompts across multi-modalities.}
    \label{fig:intro_illustration}
    \vspace{-20pt}
\end{figure}

We posit that a positive answer is achievable if the \textit{Cross-modality Alignment} capability can be pretrained using multimodal encoders~\cite{clip, li2024decisionnce, zhang2024connect, liang2022mind}.  
Consider a multimodal encoder capable of producing representations for different modalities of prompts that share an identical high-level task goal. Can we find an effective way to encode the textual and visual embeddings of prompts that describe the same task
 (e.g. ``open the door")   
in a unified representation space~\cite{huh2024platonic}? If so, prompts across different modalities will become interchangeable in this shared latent space, allowing unimodal data to implicitly serve as a proxy for multimodal data, which have been demonstrated possible in other multimodal domains like text-to-image generation and different modalities are converging in representation spaces
~\cite{huh2024platonic, zhang2024connect, li2023decap, liang2022mind}.

However, two main challenges must be addressed to apply this methodology to robots: \textit{1)} Existing multimodal encoders 
are  not directly applicable in our setting. They  either are not tailored for robot learning~\cite{clip}, or 
trained solely on narrow scopes of human activity data~\cite{goyal2017something,damen2018epick, grauman2022ego4d}, different from robotics domain~\cite{li2024decisionnce, ma2023liv, nair2022r3m, karamcheti2023voltron}. \textit{2)} Even with well-adapted multimodal encoders, a \textit{modality gap} between the representations of different modalities still persists~\cite{zhang2024connect, li2023decap, liang2022mind}, 
preventing the convergence of different modalities into a consistent latent space.


We propose \textit{Robo-MUTUAL} (\underline{Robo}tic \underline{MU}ltimodal \underline{T}ask specifications via \underline{U}nimod\underline{A}l \underline{L}earning). This new framework enhances the 
\textit{Cross-modality Alignment} capability of existing multimodal encoders by consuming a broader spectrum of robot-relevant data. Specifically, we retrain DecisionNCE~\cite{li2024decisionnce}, a state-of-the-art robotic multimodal encoder on an all-encompassing dataset, which not only consists large-scale robot datasets including Open-X~\cite{padalkar2023open} and DROID~\cite{khazatsky2024droid}, but also incorporates a large human-activity dataset EPICK-KITCHEN~\cite{damen2018epick}. Combined, these datasets 
form the most comprehensive collection  to date for robotic multimodal encoder pretraining. Building on the pretrained encoders, we explore two training-free methods to bridge the \textit{modality gap} within the representation space, where we further introduce an effective cosine-similarity noise to facilitate efficient data augmentation in representation space to enable generalization to new task prompts.
Tested across over 130 tasks and 4000 evaluations on both simulated LIBERO~\cite{liu2024libero} environments and the real robots platforms, extensive experiments showcase a promising avenue towards enabling robots to understand multimodal instructions via unimodal training.
\section{RELATED WORK}
\subsection{Task Specification in Robot Learning}
A series of task specifications have been extensively explored in robot learning. 
Some works utilize flexible specifications across various modalities to specify the tasks, such as language instructions~\cite{brohan2022rt1, brohan2023rt2, driess2023palme, octo_2023, kim2024openvla, li2024decisionnce, ma2023liv, karamcheti2023voltron, xiao2022robotic, belkhale2024rt, shi2024yell} or visual goals~\cite{ma2023vip, nasiriany2019planning, cui2022can, cui2023from, yu2018one, bonardi2020learning, james2018task}, but are specialized for unimodal task specifications.
In recent years, some works explore multimodal task specifications~\cite{lynch2020language, myers2023goal, mees2022matters, jang2022bc, shah2023mutex, jiang2023vima, yu2023using} and demonstrate positive benefits of leveraging multimodal prompts to enhance the robot policies. In these methods, \textit{Cross-modality Alignment} is one critical capability for multimodal task specifications, where multimodal prompts that share the same high-level task goal should be encoded as similar representations to encourage a well-organized representation space~\cite{shah2023mutex, mees2022matters, myers2023goal, yu2023using}. However, these methods typically train such ability from scratch on limited robot data with carefully annotated or synthetic multimodal prompts.
In this paper, instead, we aim to bypass the restrictive demands on paired multimodal prompts but directly utilize unimodal prompts to achieve multimodal task specifications utilizing the powerful pretrained mutlimodal encoders.


\subsection{Multimodal Representation in Robot Learning}

Numerous multimodal encoders, such as CLIP~\cite{clip}, BLIP~\cite{li2022blip} and BLIP2~\cite{li2023blip}, are designed to align various modalities within a unified representation space, demonstrating notable success in various areas such as image/video caption~\cite{mokady2021clipcap, hessel2021clipscore, ventura2024learning}, text-to-image generation~\cite{lafite}, and also robotics~\cite{khandelwal2022simple}. However, these encoders are not optimally suited for robot learning as they often fail to capture the temporal visual dynamics critical for robotics~\cite{ma2023vip}. To address this shortfall, specialized robotic multimodal encoders, such as R3M~\cite{nair2022r3m}, LIV~\cite{ma2023liv}, Voltron~\cite{karamcheti2023voltron}, Lorel~\cite{nair2022lorel} and the recent SOTA DecisionNCE~\cite{li2024decisionnce},  have been developed to extract these robotic-critical features. However, the \textit{Cross-modality Alignment} ability of these models is constrained by the narrow scope of their training data, typically limited to specific human datasets~\cite{damen2018epick, grauman2022ego4d, goyal2017something}, without covering diverse out-of-domain robotic data~\cite{khazatsky2024droid, walke2023bridgedata, damen2018epick, padalkar2023open}. 
\subsection{Modality Gap in Multimodal Representations}
Human can easily summarize similar mental thoughts from multimodal prompts~\cite{zhao2024eye, denervaud2020multisensory, burr2012multisensory}, which is also noticed in neural networks~\cite{huh2024platonic} where multimodal representations with same semantics are converging. This shows the potential for multimodal encoders to construct a well-aligned representation space, where modalities are interchangeable and unimodal data can approximate multimodal information. However, a persistent modality gap remains even with well-trained multimodal encoders, preventing the convergence of different modality features~\cite{liang2022mind}. Several methods such C3~\cite{zhang2024connect}, CapDec~\cite{nukrai2022capdec}, and LAFITE~\cite{lafite}, etc~\cite{schrodi2024two} try to bridge this gap, but the efficacy of these methods and how to enhance the generalization for robotics remain unexplored.

\begin{figure*}[t]
    \centering
    \includegraphics[width=1.0\linewidth]{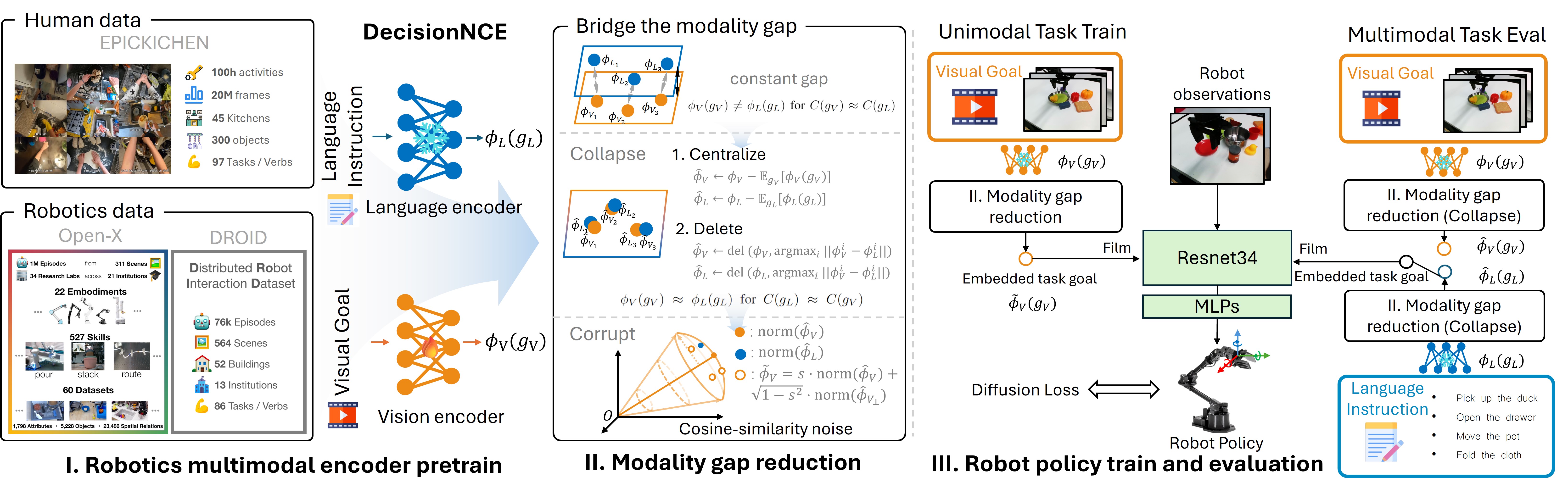}
    \vspace{-15pt}
    \caption{Robo-MUTUAL training pipeline. \textit{I}. Pretrain robotic multimodal encoder consuming broader out-of-domain human and robotics data. \textit{II}. Utilize the pretrained powerful \textit{Cross-modality Alignment} capability and further bridge the modality gap in an efficient and training-free manner. \textit{III}. Achieve multimodal task specifications via unimodal task learning leveraging the well-aligned multimodal representations.}
    \label{fig:method}
    \vspace{-5pt}
\end{figure*}

\section{Method}
\subsection{Problem Formulation}
We aim to learn a goal-conditioned policy $\pi_\theta(a|s, {g})$ that observes state $s\in\mathcal{S}$ and outputs actions $a\in\mathcal{A}$ conditioned on a task prompt ${g}$ in various modalities, using a small in-domain robot dataset $\mathcal{D}_I=\{(s_i,a_i,{g}_i)\}_{i=1}^N$. In this paper, we study the most popular prompt $g$ modalities including free-form language instruction $g_L\in\mathcal{G}_L$ and fine-grained visual goal $g_V\in\mathcal{G}_V$~\cite{lynch2020language, myers2023goal, mees2022matters, jang2022bc}, \textit{i.e.}, $g\in\{g_V, g_L\}$ and will explore more modalities like audio in future work. We assume the prompts $g$ in dataset $\mathcal{D}_I$ are unimodal.
For instance, in the case of $g_L$ is missing, we can only obtain a unimodal visual-goal conditioned policy $\pi_\theta(a|s,g_V)$ in common sense. In this paper, we also hope to let $\pi_\theta$ understand language prompts and execute similar behaviours given $g_L$ that depicts the same task as $g_V$, \textit{i.e.}, $d^{\pi_\theta}_{g_L}(s)\approx d^{\pi_\theta}_{g_V}(s)$, $\forall s\in\mathcal{S}$ for $C(g_L)\approx C(g_V)$, where $C(g)$ is the high-level task described by prompts $g$ across different modalities, $d^{\pi_\theta}_{g_L}$ and $d^{\pi_\theta}_{g_V}$ denote the state occupancy, or widely known as visitation distribution~\cite{nachum2019dualdice, li2023mind}, following the policy $\pi_\theta(a|s,g_L)$ and $\pi_\theta(a|s,g_V)$, respectively.

\subsection{Cross-modality Alignment in Representation Space}
In standard training process, mostly, $d^{\pi_\theta}_{g_L}(s)\neq d^{\pi_\theta}_{g_V}(s)$, since language instructions $g_L$ and visual goals $g_V$ are two distinct modalities, \textit{i.e.}, $g_L\neq g_V$, where the central challenge is \textit{Cross-modality Alignment}, that similar high-level tasks $C(g_L)\approx C(g_V)$ should be extracted by the policy $\pi_\theta$ to avoid this conflicts~\cite{shah2023mutex,jiang2023vima,yu2023using, nguyen2021practical, mees2022matters, jang2022bc}. 

We aim to achieve strong \textit{Cross-modality Alignment} ability by utilizing  powerful multimodal encoders $(\phi_V, \phi_L)$ pretrained on a diverse out-of-domain dataset $\mathcal{D}_{O}$ instead of training from scratch based on the limited in-domain robot data $\mathcal{D}_{I}$ like previous works~\cite{shah2023mutex, jang2022bc, mees2022matters, nguyen2021practical}. If such encoders are accessible, multimodal prompts that encode similar tasks can be projected as interchangeable representations, \textit{i.e.}, $\phi_V(g_V)\approx \phi_L(g_L)$ for $C(g_L)\approx C(g_V)$.
Then, we can train multimodal policies $\pi_\theta$ capable of understanding prompts across various modalities within this unified representation space trained solely on unimodal prompts.
\begin{equation}
\begin{aligned}
    C(g_V)\approx C(g_L) &\Leftrightarrow \phi_V(g_V)\approx\phi_L(g_L)\\ &\Leftrightarrow d^{\pi_{\theta, \phi_L}}_{g_L}(s)\approx d^{\pi_{\theta,\phi_V}}_{g_V}(s), \forall s\in \mathcal{S}.
    \label{equ:train_deploy}
\end{aligned}
\end{equation}
where $\pi_{\theta, \phi_L}$, $\pi_{\theta, \phi_V}$  denotes $\pi_\theta(a|s,\phi_L(g_L))$ and $\pi_\theta(a|s,\phi_V(g_V))$, respectively.
Targeting this goal, we propose \textit{Robo-MUTUAL} (\underline{Robo}tic \underline{MU}ltimodal \underline{T}ask specifications via \underline{U}nimod\underline{A}l \underline{L}earning), containing three parts (Fig~\ref{fig:method}): \textit{Robotic Multimodal Encoder Pretrain} (Section~\ref{subsec:mmencoder_train}), \textit{Modality Gap Reduction} (Section~\ref{subsec:modality_gap}), and \textit{Robot Policy Train and Evaluation}~\ref{subsec:robot_train_eval}).

\begin{figure}[t]
    \centering
    \includegraphics[width=1.0\linewidth]{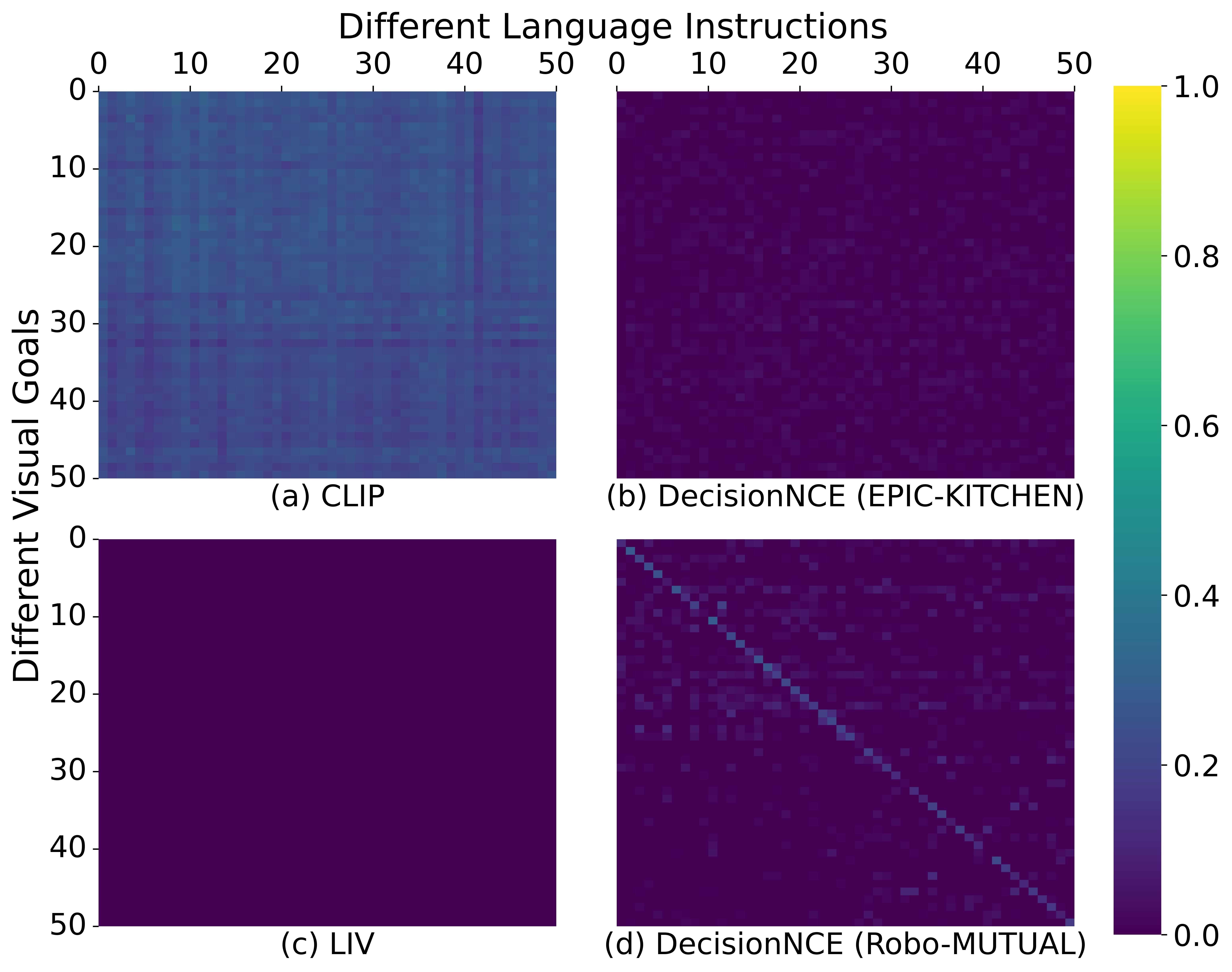}
    \vspace{-15pt}
    \caption{Heatmaps of cosine similarity between representations of language and visual goals. The diagonals are matched pairs. DecisionNCE (Robo-MUTUAL) enjoys strong \textit{Cross-modality Alignment} capability after absorbing broader out-of-domain data.}
    \vspace{-15pt}
    \label{fig:mmencoder_fail}
\end{figure}
\subsection{Robotic Multimodal Encoder Pretraining}
\label{subsec:mmencoder_train}
Numerous multimodal encoders are available ~\cite{clip, ma2023liv, li2024decisionnce}, but are not specifically designed for robotics or trained solely on limited human video that fails to cover the diverse robotics domain. We evaluate the \textit{Cross-modality Alignment} capability of several popular multimodal encoders including CLIP~\cite{clip}, LIV~\cite{ma2023liv} and DecisionNCE~\cite{li2024decisionnce}.  Fig~\ref{fig:mmencoder_fail} shows that these encoders fail to align matched visual goals and language instructions well in the robotics domain. 

Hence, we aggregate diverse large-scale robot-relevant data for robotic multimodal encoder training, aiming at improve the \textit{Cross-modality Alignment} capability as much as possible. This dataset includes Open-X dataset~\cite{padalkar2023open}, DROID~\cite{khazatsky2024droid}, and EPICK-KITCHEN~\cite{damen2018epick}, 
forming a comprehensive dataset $\mathcal{D}_O$ that spans diverse skills/tasks and scenarios. 
Based on this dataset, numerous robotic multimodal encoders could be viable, we opted to retrain the recent SOTA DecisionNCE~\cite{li2024decisionnce} for its superior temporal consistency and less sensitivity for hyper-parameter tuning. 
In difference, we freeze the language encoder $\phi_L$ from the pretrained CLIP~\cite{clip} model, focusing solely on contrasting visual goals, as we observe CLIP~\cite{clip} exhibits robust textual generalization, which is likely attributed to the extensive language data used in CLIP pretraining compared to that in robotics domains. 
We denote our encoder as DecisionNCE (Robo-MUTUAL) and the training objective is as follows:
\begin{equation}
\label{loss_objective}
\small
    \min_{\phi_V} \frac{1}{B} \sum\nolimits_{i=1}^{B}
    -\log\frac{\exp\mathcal{S}(\phi_V(o_{n_i+m_i})-\phi_V(o_{n_i}),\phi_L(l_i))}
    {\sum_{j=1}^{B} \exp\mathcal{S}(\phi_V(o_{n_j+m_j})-\phi_V(o_{n_j}),\phi_L(l_i))},
\end{equation}
where $n$ is a random selected start frame in a video clip, $m$ is a random segmentation length and $B$ is batch size. We set $B=1024$ and train it on 8$\times$A100 GPU for 8 days.
See from Fig~\ref{fig:mmencoder_fail} that DecisionNCE (Robo-MUTUAL) aligns matched visual and textual goals well, where $\mathcal{S}\left(\phi_V(g_V), \phi_L(g_L)\right)$ is high for $C(g_V)\approx C(g_L)$ 
and $\mathcal{S}$ is cosine similarity. Considering this superior ability,  We will release our check point for DecisionNCE (Robo-MUTUAL) to support researchers develop other future applications conveniently.
\begin{figure}[t]
    \centering
    \includegraphics[width=1.0\linewidth]{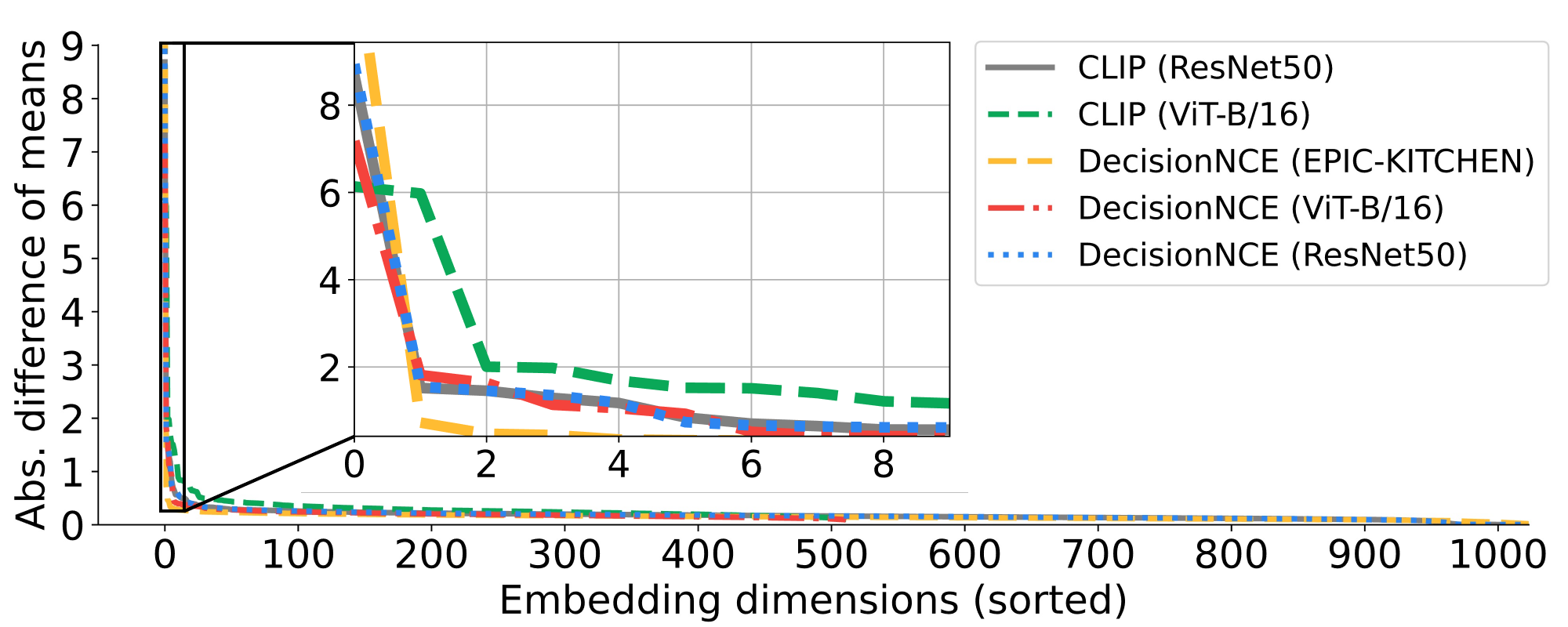}
    \vspace{-15pt}
    \caption{Abs. difference in the means of each embedding dimension cross different modalities. The modality gap manifests in a few dimensions with large discrepancies across modalities, while others remain consistent.}
    \label{fig:diff_plt}
    \centering
    \vspace{+2pt}
    \includegraphics[width=1.0\linewidth]{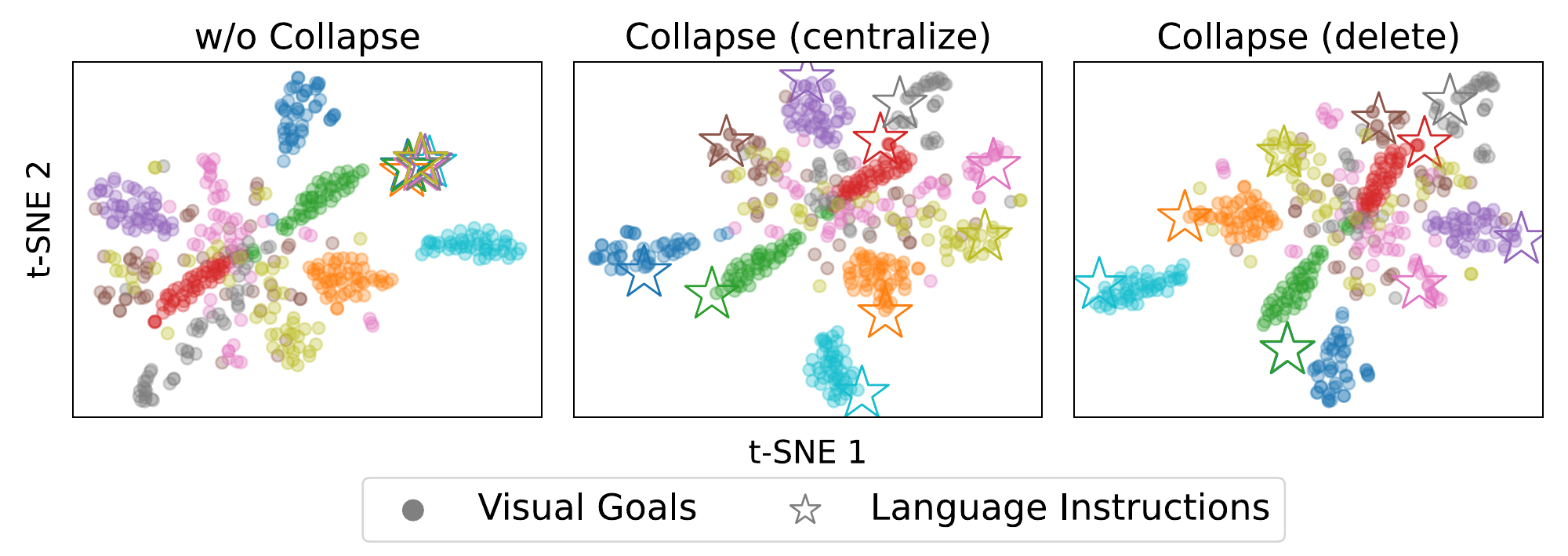}
    \vspace{-15pt}
     \caption{t-SNE~\cite{van2008visualizing} projection of DecisionNCE (Robo-MUTUAL) representations of matched visual and language goals. Although different modalities are separate initially, indicating a huge modality gap, this gap can be reduced through the simple \textit{centralize} or \textit{delete} Collapse methods.}
    \label{fig:modality_gap}
    \vspace{-10pt}
\end{figure}

\subsection{Modality Gap Reduction}
\label{subsec:modality_gap}
Nevertheless, huge modality gap exists in the pretrained representation space, where different modalities with similar semantic meanings fail to absolutely converge but instead are clustered per modality, resulting in $\phi_V(g_V)\neq\phi_L(g_L)$ for $C(g_V)\approx C(g_L)$ 
~\cite{zhang2024connect, jing2022understanding}, as shown in Fig~\ref{fig:modality_gap}.
Inspired by recent progresses in minimizing the modality gaps~\cite{zhang2024connect, li2023decap, liang2022mind, zhang2023diagnosing}, we apply two simple yet efficient \textit{Collapse} and \textit{Corrupt} manipulation to fill the gap.


\noindent \textbf{Collapse.} \ Specifically, the modality gap is primarily characterized as a constant gap between the span of the visual and textual representations clusters $e_V:=\{\phi_V(g_{V})\},g_V\sim\mathcal{G}_V$ and $e_L:=\{\phi_L(g_{L})\}, g_L\sim\mathcal{G}_L$, as proved in~\cite{zhang2024connect, liang2022mind}. This gap manifests in a few dimensions of the embeddings, which display significant discrepancies across modalities, while the remaining dimensions remain consistent, as shown in~\cite{schrodi2024two} and is also observed in Fig~\ref{fig:diff_plt}. Thus, intuitively this gap can be straightforwardly removed 
by deleting the most different dimensions to address the modality gap~\cite{schrodi2024two}.
\begin{equation}
    \begin{aligned}
             {\rm detele:} \ &\hat{\phi_V}\leftarrow {\rm del} (\phi_V, \arg\max_i \|\phi_V^i-\phi_L^i\|), \\ &\hat{\phi_L}\leftarrow {\rm del} (\phi_L, \arg\max_i \|\phi_V^i-\phi_L^i\|),
    \end{aligned}
\end{equation}
where, $\phi^i$ denotes the $i$-th dimension of $\phi$, and ${\rm del}(\phi, i)$ denotes deleting the $i$-th dimension of $\phi$. Or we can reduce each modality by its mean values to consider the modality difference across all dimensions, as proved in~\cite{zhang2023diagnosing, liang2022mind, zhang2024connect}:
\begin{equation}
\begin{aligned}
     {\rm centralize:} \ &\hat{\phi_V}\leftarrow \phi_V - \mathbb{E}_{g_V}\left[\phi_V(g_V)\right],\\ &\hat{\phi_L}\leftarrow \phi_L - \mathbb{E}_{g_L}\left[\phi_L(g_L)\right],
\end{aligned}
\end{equation}
where we approximate the expectation term via mini-batch samples from the dataset $\mathcal{D}_O$. After these simple manipulation, the collapsed multimodal representations can mostly bridge the modality gap and collapse together, \textit{i.e.}, $\hat{\phi_V}(g_V)\approx\hat{\phi_L}(g_L)$ for $C(g_V)\approx C(g_L)$, as shown in Fig~\ref{fig:modality_gap}. In our paper, we report the main results using \textit{Centralize} by default and provide ablation studies in experiments.

\noindent \textbf{Corrupt.} \ 
Note that one language can describe diverse visual goals and vice versa. Hence, we can conveniently augment the collapsed representations to improve the generalization by adding some random noise, which enables $\pi_\theta$ potentially understand unseen prompts. Some works argue that the simple Gaussian noise is effective~\cite{zhang2024connect}, but we found the cosine similarity noise offers superior augmentation in the high-dimensional embedding spaces~\cite{liu2024arcsin}. Here we slightly abuse the notation by simplifying $\phi_V$ or $\phi_L$ as $\phi$:
\begin{equation}
\label{cosine}
    \begin{aligned}
    \small
        \tilde{\phi}\leftarrow s \cdot{\rm norm}(\hat{\phi})+\sqrt{1-s^2}\cdot {\rm norm}({\phi}_{\perp}), \hat{\phi}_{\perp} = {v-\frac{v\cdot\hat{\phi}}{\hat{\phi}\cdot \hat{\phi}}\cdot \hat{\phi}}
    \end{aligned}
\end{equation}
where $v$ is a random embedding, $\hat{\phi}_{\perp}$ is the orthogonal vector in $v$ \textit{w.r.t} $\hat{\phi}$, 
$\rm norm(x):=\frac{x}{\|x\|}$, 
$s$ is a cosine similarity randomly selected from $[\alpha, 1]$ and we set $\alpha$ to 0.2 as default. 
The direction of augmented representations remain close to the original to preserve the semantics after corruption, \textit{i.e.}, in (\ref{cosine}), $\mathcal{S}(\tilde{\phi}, \hat{\phi})=\frac{\tilde{\phi}\cdot\hat{\phi}}{\|\tilde{\phi}\|\|\hat{\phi}\|}=s\ge\alpha$, which however can be potentially destroyed by a too large Gaussian noise.


\subsection{Robot Policy Training and Evaluation}
\vspace{-2pt}
\label{subsec:robot_train_eval}
Now, we can achieve \textit{Cross-modality Alignment} in representation space. For instance, in scenarios where only visual goals $g_V$ are available and language instructions $g_L$ are absent, we can utilize the pretrained DecisionNCE (Robo-MUTUAL) visual encoder $\phi_V$ to generate the visual goal representations $\phi_V(g_V)$, then derive the final corrupted representations $\tilde{\phi_V}(g_V)$ for training the robot policy $\pi_\theta$. During deployment, the policy $\pi_\theta$ can be prompted with either visual goals $g_V$ or language goals $g_L$, by using the collapsed and aligned representations $\hat{\phi_V}(g_V)$ or $\hat{\phi_L}(g_L)$.

In training details, we utilize ResNet34~\cite{He_2016_CVPR} to extract visual feature from both a base and wrist view, where task embedding $\phi$ is injected via Film conditioning layers~\cite{perez2018film}. Then, the visual feature is passed through a residual MLPs to predict actions similar to IDQL~\cite{hansen2023idql}. The policy is optimized with diffusion loss~\cite{ho2020denoising} for its superior effectiveness to model complex distributions~\cite{chi2023diffusion, zheng2024safe, walke2023bridgedata}. We also use action chunking~\cite{zhao2023learning} to improve the policy smoothness.

%







\section{Experiments}
\begin{figure}[h]
    \centering
    \includegraphics[width=1.0\linewidth]{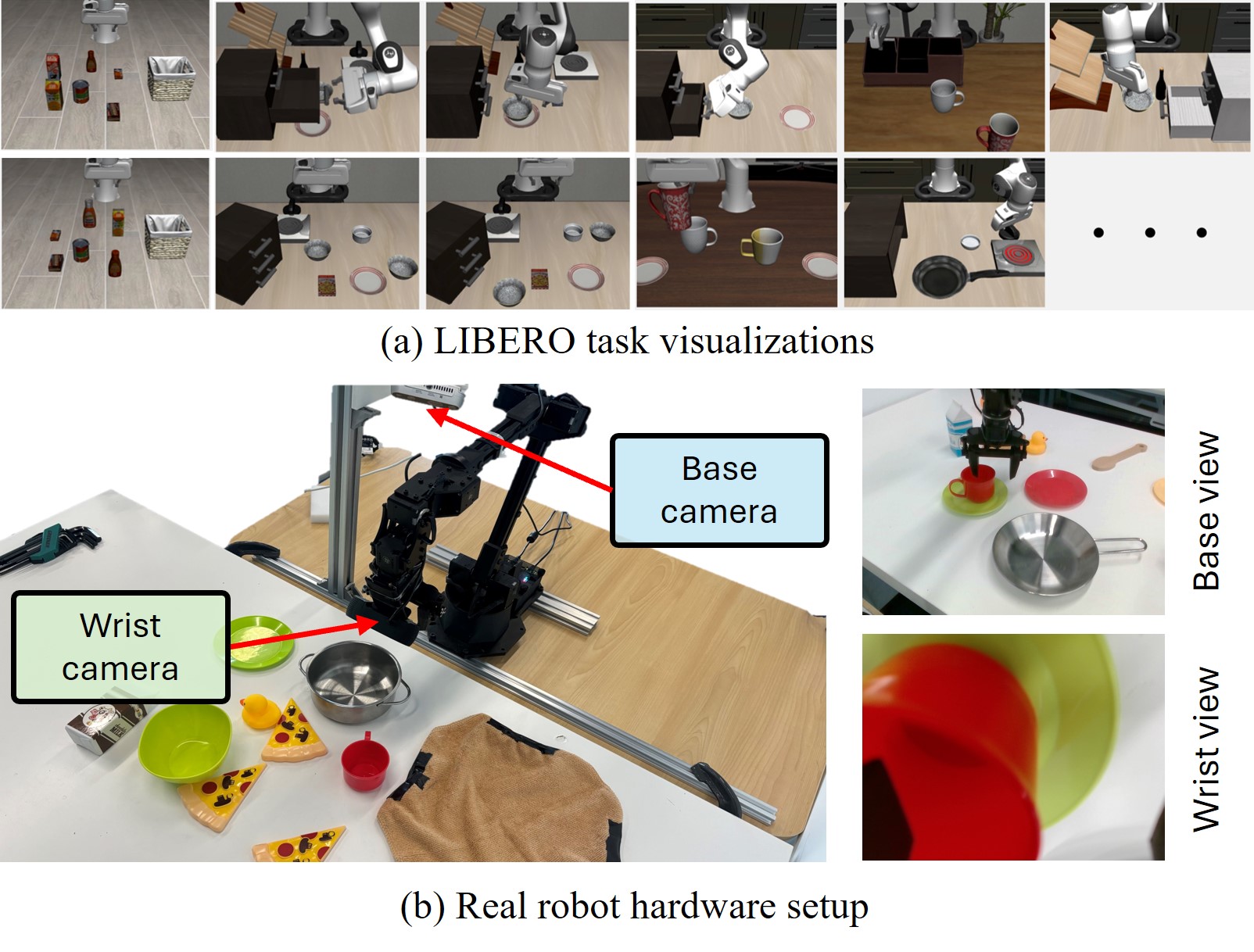}
    \vspace{-20pt}
    \caption{Simulation and real world robotics evaluation setups.}
    \vspace{-10pt}
    \label{fig:setup}
\end{figure}

\begin{figure*}[h]
    \centering
    \includegraphics[width=1.0\linewidth]{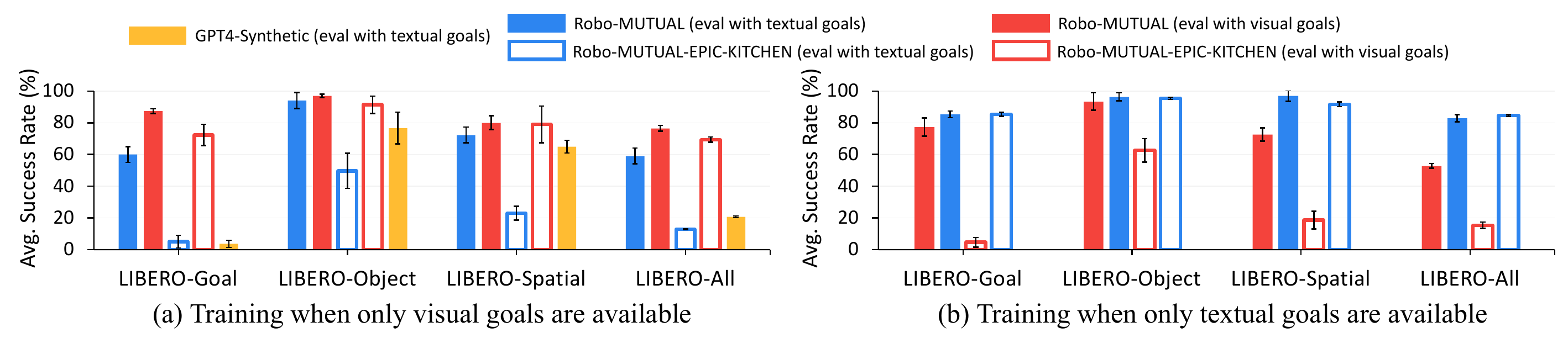}
    \vspace{-18pt}
    \caption{\textbf{Simulation evaluation.} (eval with textual/visual goals) denote the robot policy is evaluated with textual goal and visual goals, respectively. For each bar, the robot policy is trained on 3 different random seeds and is evaluated for 10 episodes for each task.
    }
    \label{fig:exp_sim}
\vspace{+2pt}
    \includegraphics[width=1.0\linewidth]{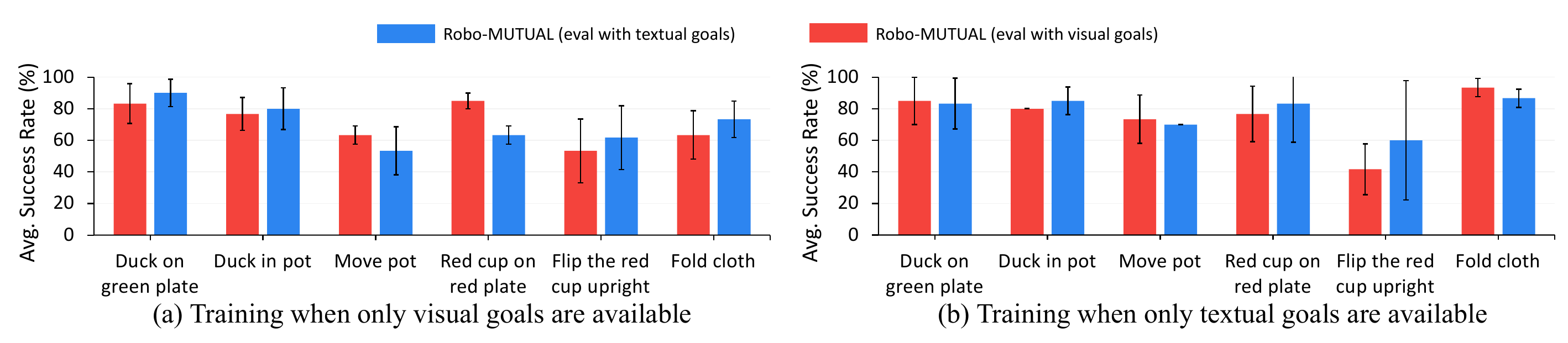}
    \vspace{-18pt}
    \caption{\textbf{Real robot evaluation.} (eval with textual/visual goals) denote the robot policy is evaluated with textual goal and visual goals, respectively. For each bar, the robot policy is trained on 3 different random seeds and is evaluated for 10 episodes for each task.}
    \vspace{0pt}
    \label{fig:exp_real}
\end{figure*}

The experiments try to answer the following questions:
\begin{itemize}
    \item Can Robo-MUTUAL achieve multimodal task specifications with unimodal data training?
    \item Does Robo-MUTUAL outperform baseline models that use synthetic prompts generated from unimodal data for multimodal task specifications?
    \item How do specific design choices for Robo-MUTUAL, such as the use of enhanced robotic multimodal encoders, various collapse, corruption methods, and different scales of corruption, contribute to its effectiveness?
\end{itemize}
\subsection{Experimental Setup}
\noindent\textbf{Simulation Environments}. We employ the LIBERO benchmark \cite{liu2024libero}, which consists of 130 robotic simulation tasks across four distinct suites: LIBERO-Goal/Object/Spatial/100. LIBERO-100 contains 100 tasks, while the other three suites each have 10 tasks. Each task is accompanied by a language instruction $g_L$ and 50 expert trajectories. Collectively, these suites contain 6500 expert demonstrations and 130 different tasks, which we refer to as LIBERO-All in this paper.

\noindent\textbf{Real Robot Environments.} We evaluate 6 tasks on a WidowX robot, spanning skills include \texttt{pick \& place}, \texttt{fold}, \texttt{flip} and \texttt{move}. We collect around 100 demonstrations per task using the Bridgedata system~\cite{walke2023bridgedata}, and annotate each task with a language instruction $g_L$. Successful catching correct object is recorded as half-completed of the whole task, and a continued correct placement will scored a full success.

\noindent\textbf{Evaluation Scenarios}. \textit{1)} We train Robo-MUTUAL in a common scenario where only visual goals $g_V$ are available. Visual goal is provided as the transition between the representations of the initial and final frames of a video clip~\cite{li2024decisionnce}. In this case, 
the policy only observes visual goals $g_V$ during training but is required to understand both visual goals $g_V$ and language instructions $g_L$ during evaluation. \textit{2)} We also explore the reverse scenario, \textit{i.e.}, training exclusively on textual goals $g_L$ but evaluating with visual goals $g_V$ to assess bidirectional transferability across modalities. Due to the high cost associated with real-world evaluations, we compare Robo-MUTUAL against baseline methods in simulations. However, we believe the 130 tasks in LIBERO benchmark offer comprehensive coverage across a diverse range of tasks and skills, ensuring fair and sufficient comparisons.


\noindent\textbf{Baselines.} 
    \textbf{1) GPT4-Synthetic} is the main baseline in the common scenario that textual goals are missing. This approach represents a series of methods that firstly generate missing language instructions from unimodal visual goals using pretrained large multimodal models~\cite{xiao2022dial,blank2024scaling}, and then train policies conditioned on these synthetic goals. During evaluation, we prompt the policies using the ground truth language instructions to evaluate the success rates. To ensure a fair comparison, we employ the advanced GPT-4V to generate language instructions based on the initial and final frames of a trajectory. Furthermore, we incorporate several ground truth examples to enhance the quality of synthetic instructions leveraging the substantial in-context learning capability of large models~\cite{dong2022survey}. 
    \textbf{2) Robo-MUTUAL-EPICK} adheres the same implementation protocols as our Robo-MUTUAL, but directly uses the DecisionNCE~\cite{li2024decisionnce} pretrained solely on the narrow EPICK-KITCHEN dataset~\cite{damen2018epick} as the robotic multimodal encoder without the specialized improvement in Section~\ref{subsec:mmencoder_train}, which may suffer from limited \textit{Cross-modality Alignment} capability. 

\begin{figure*}[t]
    \includegraphics[width=1.0\linewidth]{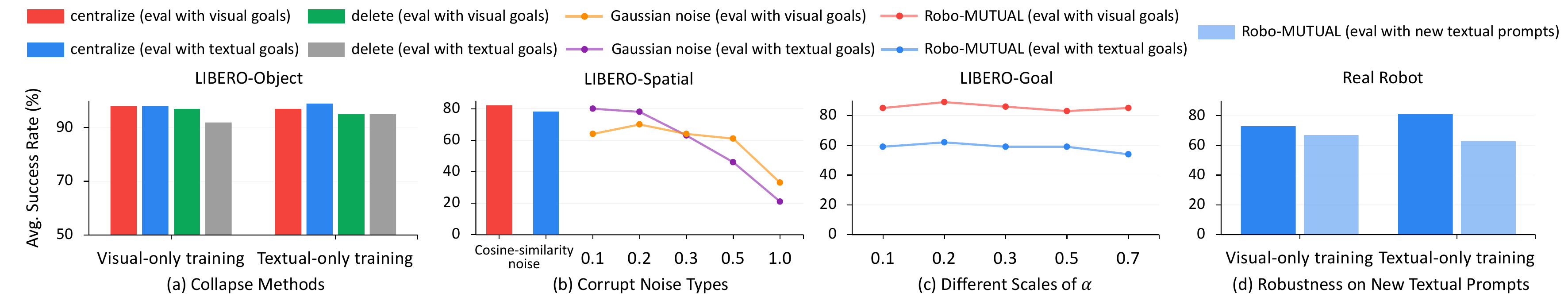}
    \vspace{-15pt}
    \caption{\textbf{Ablation studies.} (a) Comparison between (\textit{delete} and \textit{centralize}) collapse methods; (b) Comparison between \textit{Cosine-similarity noise} and Gaussian noise; (c) Corrupt strength of \textit{Cosine-similarity noise}; (d) The generalization ability on unseen new textual instructions}
    \vspace{-10pt}
    \label{fig:ablation}
\end{figure*}

\vspace{-3pt}
\subsection{Main Results}
\vspace{-3pt}
\label{subsec:train multi via uni}
\noindent\textbf{Transfer from Visual to Textual Goals}. 
In the common scenario where only visual goals are available, both the simulated and real-world evaluations in Fig~\ref{fig:exp_sim} (a) and Fig~\ref{fig:exp_real} (a) clearly demonstrate that Robo-MUTUAL can successfully understand both visual and textual goals trained exclusively on visual goals. Specifically, we only observe a moderate performance drop when evaluating Robo-MUTUAL from visual to textual goals. However, this property is not enjoyed by Robo-MUTUAL-EPICK, which fails to transfer from visual to textual goals and undergoes severe performance drop, as shown in Fig~\ref{fig:exp_sim} (a), primarily due to the limited \textit{Cross-modality Alignment} capability (see Fig~\ref{fig:mmencoder_fail} for details). Moreover, GPT4-Synthetic can partially work well using the synthetic language instructions thanks to the superior capability of large models, however, still underperforms Robo-MUTUAL.
We think this is because GPT-4V only consider limited robotic data during pretraining and it is hard to guarantee the quality of synthetic data~\cite{xiao2022dial}.

\noindent\textbf{Transfer from Textual to Visual Goals}. We also explore the reverse scenario in which Robo-MUTUAL is trained exclusively on textual goals and then evaluated with visual goals. In Fig~\ref{fig:exp_sim} (b) and Fig~\ref{fig:exp_real} (b), Robo-MUTUAL demonstrates effective transferability in this reversed setup. Importantly, the visual goals provided for task specifications do not align precisely with the robot's current observations, \textit{i.e.}, discrepancies exist in the location of target object, the target position, and even the number, location, types of distractors. Under this hard condition, Robo-MUTUAL must accurately extract correct semantic task information from visual goals that contain many distracting elements in a zero-shot manner trained solely with textual goals. Overall, Robot-MUTUAL demonstrates strong bidirectional transferability to enable multimodal task instructions via unimodal learning\footnote{See detailed videos in \href{zh1hao.wang/Robo_MUTUAL}{\texttt{zh1hao.wang/Robo\_MUTUAL}}.}. 

\subsection{Ablation Studies}
\label{subsec:ablation}
\noindent\textbf{{Collapse} Methods}. 
In the main results, we only report \textit{centralize} due to space limits. Here, we compare the \textit{centralize} and \textit{delete} collapse methods. Fig~\ref{fig:ablation}(a) shows that they achieve comparable performance and enjoy moderate performance gap. Also, Fig~\ref{fig:modality_gap} shows that both of them effectively cluster matched textual and visual pairs. One future work is to investigate the conditions under which \textit{centralize} and \textit{delete} outperform each other in the robotics domain.



\noindent\textbf{{Corrupt} Noise Types.} We compare our \textit{Cosine-similarity noise} with the widely adopted simple Gaussian noise with different standard deviation (Std)~\cite{zhang2024connect}. Fig~\ref{fig:ablation} (b) demonstrates that the simple Gaussian noise is quite unstable \textit{w.r.t} different Std and is inferior to \textit{Cosine-similarity noise}. Intuitively, this is because \textit{Cosine-similarity noise} naturally enables the augmented data to preserve the semantics of its origins, as the augmented data remains a high cosine similarity with the original data, \textit{i.e.}, $\mathcal{S}(\tilde{\phi}, \hat{\phi})\in[\alpha,1]$. However, this can be easily violated by the simple Gaussian noise, as a too large Gaussian noise can mostly reverse the direction of the original representation after augmentation and thus lost its original semantics. On the other hand, however, a too small noise lacks enough augmentations, making it quite sensitive to tune the Std of the Gaussian noise.


\noindent\textbf{{Corrupt} Strength}. To further demonstrate the effectiveness of \textit{Cosine-similarity noise}, we ablate on different corrupt strength by adjusting the cosine-similarity threshold $\alpha$. Fig~\ref{fig:ablation} (c) shows that \textit{Cosine-similarity noise} is robust to various corrupt strength, maintaining consistent performance across various $\alpha$. Meanwhile, $\alpha=0.2$ works best in our evaluation and is set as our default choice. This likely correlates with the pre-corruption cosine-similarity of matched pairs, which averages around 0.2, as shown in Fig~\ref{fig:mmencoder_fail}(d).

\noindent\textbf{Robustness on New Textual Prompts}. We also investigate whether Robot-MUTUAL can generalize beyond the ground-truth goals in the downstream dataset $\mathcal{D}_I$. For example, we replace the original ``pick up the red cup and place it on the plate" as ``move the red cup to the plate", sharing the same abstracted task goal but are expressed differently. Fig~\ref{fig:ablation} (d) shows that Robo-MUTUAL enjoys robustness to such new textual prompts thanks to the strong \textit{Cross-modality Alignment} capability that maps different instructions with similar semantics as similar representations. This means that we can conveniently augment instructions in the latent space in one implicit way by simply adding noise, like the \textit{Cosine-similarity noise}, rather than relying on heavy large models to explicitly synthesize or refine instructions like GPT4-Synthetic and other relevant works~\cite{xiao2022dial, blank2024scaling}.

\section{Conclusions}

We introduce Robo-MUTUAL, a framework that enbales robots to comprehend multimodal task specifications using only unimodal prompts. This is achieved by treating multimodal instructions as interchangeable embeddings within a well-aligned multimodal representation space, leveraging the strong \textit{Cross-modality Alignment} capability from pretrained encoders on a comprehensive robotic dataset and two simple yet effective modality gap reduction methods. Extensive evaluations on both real and simulated robots validate the effectiveness of our approach. One limitation is we only consider the language and image modalities, and we will explore more modalities like audio in future work. Further enhancements will also focus on refining multimodal encoders and improving modality gap reduction techniques.






%



\newpage
\bibliographystyle{ieeetr}
\bibliography{main}

\end{document}